\documentclass{article} 

\usepackage{iclr2024_conference,times}

\usepackage{amsmath,amsfonts,bm}

\def\eqref#1{equation~\ref{#1}}

\def\1{\bm{1}}

\DeclareMathAlphabet{\mathsfit}{\encodingdefault}{\sfdefault}{m}{sl}
\SetMathAlphabet{\mathsfit}{bold}{\encodingdefault}{\sfdefault}{bx}{n}

\usepackage{hyperref}
\usepackage{url}

\usepackage{CJKutf8}
\usepackage{arydshln}
\usepackage{booktabs}
\usepackage{colortbl}
\usepackage{multirow}
\usepackage{enumitem}
\usepackage{tcolorbox}

\newcommand{\ie}{\textit{i.e.}}
\newcommand{\eg}{\textit{e.g.}}

\renewcommand{\cite}{\citep}

\title{Revisiting the Reliability of Psychological Scales on Large Language Models}

\author{
Jen-tse Huang$^{1,2}$, Wenxiang Jiao$^2$, Man Ho Lam$^1$, Eric John Li$^1$, Wenxuan Wang$^{1,2}$\thanks{Wenxuan Wang is the corresponding author.}, Michael R. Lyu$^1$ \\
$^1$Department of Computer Science and Engineering, The Chinese University of Hong Kong \\
\texttt{\{jthuang,wxwang,lyu\}@cse.cuhk.edu.hk} \ \ \ \ \texttt{\{mhlam,ejli\}@link.cuhk.edu.hk} \\
$^2$Tencent AI Lab \ \ \ \ \texttt{\{joelwxjiao\}@tencent.com} \\
}

\iclrfinalcopy 
\begin{document}

\maketitle

\begin{figure}[h!]
    \centering
    \includegraphics[width=0.7\linewidth]{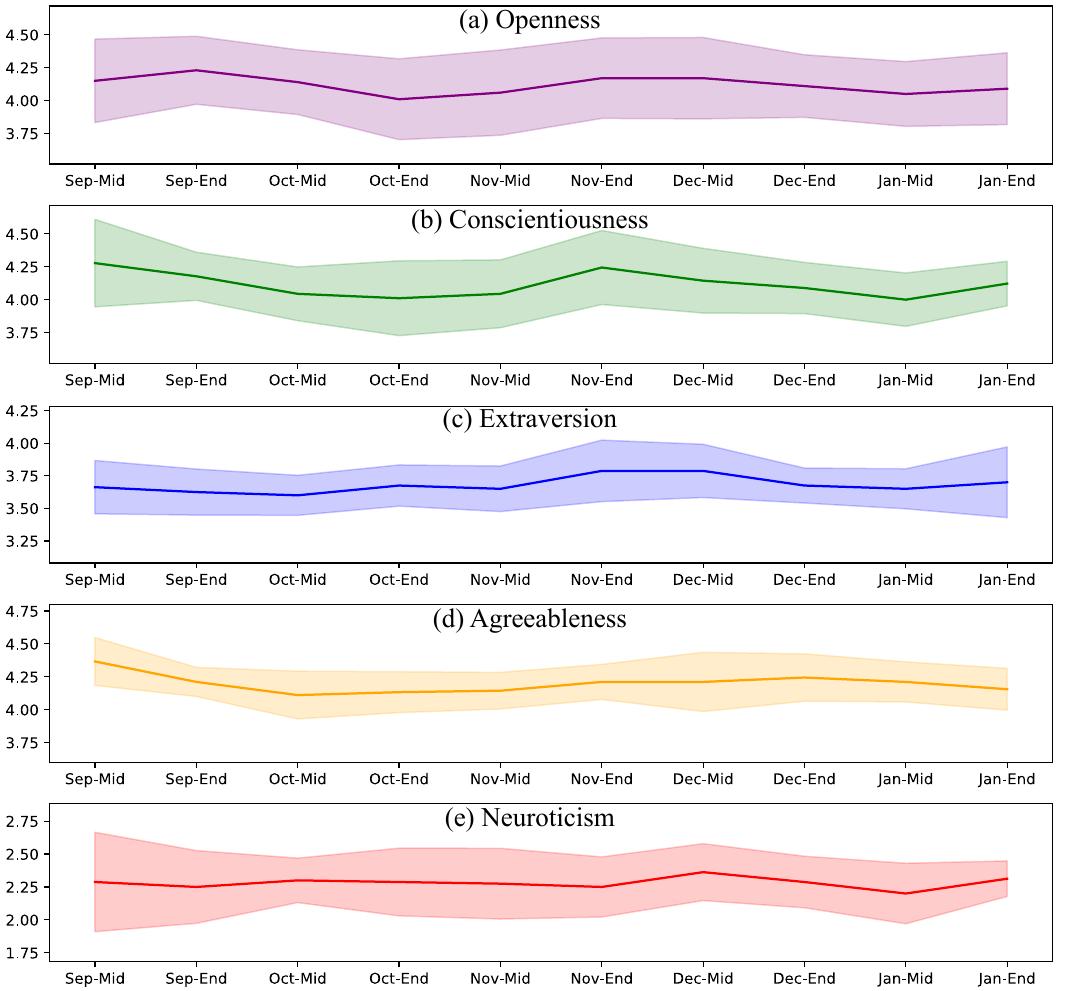}
    \caption{Biweekly measurements starting from mid-September 2023 to late-January 2024 of the BFI on GPT-3.5-Turbo. The model experienced two different versions (0613, 1106) during this period. The shadow represents the standard deviation ($\pm Std$).}
    \label{fig:test-retest}
\end{figure}

\begin{abstract}
Recent research has focused on examining Large Language Models' (LLMs) characteristics from a psychological standpoint, acknowledging the necessity of understanding their behavioral characteristics.
The administration of personality tests to LLMs has emerged as a noteworthy area in this context.
However, the suitability of employing psychological scales, initially devised for humans, on LLMs is a matter of ongoing debate.
Our study aims to determine the reliability of applying personality assessments to LLMs, explicitly investigating whether LLMs demonstrate consistent personality traits.
Analysis of 2,500 settings per model, including GPT-3.5, GPT-4, Gemini-Pro, and LLaMA-3.1, reveals that various LLMs show consistency in responses to the Big Five Inventory, indicating a satisfactory level of reliability.
Furthermore, our research explores the potential of GPT-3.5 to emulate diverse personalities and represent various groups—a capability increasingly sought after in social sciences for substituting human participants with LLMs to reduce costs.
Our findings reveal that LLMs have the potential to represent different personalities with specific prompt instructions.
\end{abstract}

\section{Introduction}

The recent emergence of Large Language Models (LLMs) marks a significant advancement in the field of Artificial Intelligence (AI), showcasing its abilities in various natural language processing tasks, including text translation~\cite{jiao2023chatgpt}, sentence revision~\cite{wu2023chatgpt}, program repair~\cite{fan2023automated}, and program testing~\cite{deng2023large}.
Furthermore, LLM applications extend beyond computer science, enhancing fields such as clinical medicine~\cite{cascella2023evaluating}, legal advice~\cite{deroy2023ready}, and education~\cite{dai2023can}.
Currently, LLMs are catalyzing a paradigm shift in human-computer interaction, revolutionizing how individuals engage with computational systems.
With the integration of LLMs, computers have transcended their traditional role as tools to become assistants, establishing a symbiotic relationship with users.
Thus, the focus of research extends beyond assessing LLM performance to understanding their behaviors from a psychological perspective.
\citet{huang2023humanity} highlights the significance of psychological analysis on LLMs in developing AI assistants that are more human-like, empathetic, and engaging.
Such analysis also plays a crucial role in identifying potential biases or harmful behaviors through the understanding of the decision-making processes of LLMs.

In this context, personality tests aimed at quantifying individual characteristics have gained popularity recently~\cite{safdari2023personality, bodroza2023personality, huang2023humanity}.
However, the applicability of psychological scales, initially designed for humans, to LLMs has been contested.
Critics argue that LLMs lack consistent and stable personalities, challenging the direct transfer of these scales to AI agents~\cite{song2023have, gupta2023investigating, shu2023you}.
The essence of this debate lies in the \textbf{reliability} of these scales when applied to LLMs.
``Reliability'' in psychological terms refers to the consistency and stability of results derived from a psychological scale.
Evaluating reliability in LLMs differs from its assessment in humans since LLMs demonstrate a heightened sensitivity to input variations compared to humans.
For example, humans generally provide consistent responses to questions regardless of their order, while LLMs might yield different answers due to varied contextual inputs.
Although consistent results can be obtained from an LLM by querying single items with a zero-temperature parameter setting, such responses are likely to vary under different input conditions.
Therefore, our study first systematically investigates the reliability of LLMs on psychological scales under varying conditions, including instruction templates, item rephrasing, language, choice labeling, and choice order.
Through analyzing the distribution of all 2,500 settings, we find that various LLMs demonstrate sufficient reliability on the Big Five Inventory.

Additionally, our study further explores whether instructions or contexts can influence the distribution of personality results.
We seek to answer whether LLMs can replicate responses of diverse human populations, a capability increasingly sought after by social scientists for substituting human participants in user studies~\cite{dillion2023can}.
However, this topic remains controversial~\cite{harding2023ai}, warranting thorough investigation.
In particular, we employ three approaches to affecting the personalities of LLMs, from low directive to high directive: (1) by creating a specific environment, (2) by assigning a predetermined personality, and (3) by embodying a character.
Firstly, recent research by \citet{coda2023inducing} demonstrates the impact of a sad/happy context on LLMs' anxiety levels.
Following this work, we conduct experiments to assess LLM's personality within these varied emotional contexts.
Secondly, we assign a specific personality for LLM, drawing upon existing literature that focuses on changing the values of LLMs~\cite{santurkar2023whose}.
Thirdly, inspired by \citet{deshpande2023toxicity}, which investigates the assignment of a persona to ChatGPT for assessing its tendency towards offensive language and bias, we instruct the LLM to embody the characteristics of a predefined character and measure the resulting personality.
Our findings indicate that GPT-3.5-Turbo can represent various personalities in response to specific prompt adjustments.

The contributions of this study are as follows:
\begin{itemize}[leftmargin=*]
    \item This study is the first to conduct a comprehensive analysis through five distinct factors on the reliability of psychological scales applied to LLMs, showing that GPT-3.5-Turbo has stable and distinct personalities.
    \item Our research contributes to the field of social science by demonstrating the potential of LLMs to simulate diverse human populations accurately.
    \item We have developed a framework for assessing the reliability of psychological scales on LLMs, which paves the way for future research to validate a broader range of scales on various LLMs.
\end{itemize}
We have made our experimental results and the corresponding code available to the public on GitHub,\footnote{\url{https://github.com/CUHK-ARISE/LLMPersonality}} promoting transparency and facilitating further research in this domain.
\section{Preliminaries}

\subsection{Personality Tests}

Personality tests are instruments designed to quantify an individual's character, behavior, thoughts, and emotions.
A prominent model for assessing personality is the five-factor model, \textit{OCEAN} (Openness, Conscientiousness, Extraversion, Agreeableness, Neuroticism), also known as the Big Five personality traits~\cite{john1999big}.
Other notable models include the Myers-Briggs Type Indicator~(MBTI)~\cite{myers1962myers} and the Eysenck Personality Questionnaire~(EPQ)~\cite{eysenck1985revised}, each based on distinct trait theories.
Extensive research has demonstrated these models' effectiveness (\ie, reliability and validity) in human subjects.
However, the application of these tests to LLMs remains a topic of debate.

\subsection{Reliability and Validity of Scales}
\label{sec:preliminaries}

\begin{table*}[t]
    \centering
    \resizebox{1.0\linewidth}{!}{
    \begin{tabular}{l p{15cm}}
    \toprule
    \bf Template & \multicolumn{1}{c}{\bf Details} \\
    \hline
    T1 \cite{huang2023humanity} & You can only reply from \texttt{START} to \texttt{END} in the following statements. Here are a number of characteristics that may or may not apply to you. Please indicate the extent to which you agree or disagree with that statement. \texttt{LEVEL\_DETAILS} Here are the statements, score them one by one: \texttt{ITEMS} \\
    \hline
    T2 \cite{miotto2022gpt} & Now I will briefly describe some people. Please read each description and tell me how much each person is like you. Write your response using the following scale: \texttt{LEVEL\_DETAILS} Please answer the statement, even if you are not completely sure of your response. \texttt{ITEMS} \\
    \hline
    T3 \cite{jiang2023evaluating} & Given the following statements of you: \texttt{ITEMS} Please choose from the following options to identify how accurately this statement describes you. \texttt{LEVEL\_DETAILS} \\
    \hline
    T4 \cite{safdari2023personality} & Here are a number of characteristics that may or may not apply to you. Please rate your level of agreement on a scale from \texttt{START} to \texttt{END}. \texttt{LEVEL\_DETAILS} Here are the statements, score them one by one: \texttt{ITEMS} \\
    \hline
    T5 \cite{safdari2023personality} & Here are a number of characteristics that may or may not apply to you. Please rate how much you agree on a scale from \texttt{START} to \texttt{END}. \texttt{LEVEL\_DETAILS} Here are the statements, score them one by one: \texttt{ITEMS} \\
    \bottomrule
    \end{tabular}
    }
    \caption{Five different versions of instructions to complete the personality tests for LLMs from different papers.}
    \label{tab-prompts}
\end{table*}

In psychometrics, the concepts of reliability and validity are crucial for evaluating the quality and effectiveness of psychological scales and tests.
\textbf{Reliability} refers to the consistency and stability of the results obtained from a psychological test or scale.
There are various types of reliability; two common ones are \textit{Test-Retest Reliability} and \textit{Internal Consistency Reliability}.
\textit{Test-Retest Reliability} assesses the stability of a test over time~\cite{guttman1945basis} while \textit{Internal Consistency Reliability} checks how well the items within a test measure the same concept or construct~\cite{cronbach1951coefficient}.
\textbf{Validity} is how well a test measures what it should measure.
Researchers usually consider different types of validity, such as \textit{Construct Validity} and \textit{Criterion Validity}~\cite{safdari2023personality}.
Being the most critical type of validity, \textit{Construct Validity} refers to how well a scale measures the theoretical construct it is supposed to measure.
\textit{Construct validity} is often demonstrated through correlations with other measures that are theoretically related (\textit{Convergent Validity}) and not correlated with measures that are theoretically unrelated (\textit{Divergent Validity})~\cite{messick1998test}.
\textit{Criterion Validity} assesses how well one measure predicts an outcome based on another measure~\cite{clark2019constructing}.
It is often split into \textit{Concurrent Validity}, when the scale is compared to an outcome that is already known at the same time the scale is administered; and \textit{Predictive Validity} when the scale is used to predict a future outcome~\cite{barrett1981concurrent}.
While reliability is a necessary but insufficient condition for validity, validity inherently necessitates reliability.
Consequently, assessing the reliability of scales forms the foundational step in evaluating the personality traits of LLMs and thus constitutes the primary focus of this study.
\section{The Reliability of Scales on LLMs}

This section focuses on evaluating the reliability of psychological scales applied to LLMs.
We first introduce the framework established for assessing the stability of responses generated by LLMs.
Subsequently, we show the findings, including both visual and quantitative data.

\subsection{Framework Design}
\label{sec:framework}

The consistency of responses from LLMs is predominantly determined by their input~\cite{hagendorff2023machine}.
To assess the reliability of LLMs, it is crucial to examine their responses across varying input conditions.
In this study, we propose to deconstruct a query into five distinct factors for a comprehensive analysis: (1) the nature of the instruction, (2) the specific items in the scale, (3) the language used, (4) the labeling of choices, and (5) the order in which these choices are presented.

\paragraph{(1) Instruction}
Given that LLMs exhibit sensitivity to variations in prompt phrasing, as observed by \citet{bubeck2023sparks}, and \citet{gupta2023investigating} highlighted that LLMs demonstrate differing personalities under varying prompting instructions, we need to evaluate the influence of different instructions.
To this end, we analyze the performance of five distinct prompt templates: T1 as applied in \citet{huang2023humanity}, T2 as used by \citet{miotto2022gpt}, T3 suggested by \citet{jiang2023evaluating}, and T4 and T5 both identified in \citet{safdari2023personality}.
Details of prompts are listed in Table~\ref{tab-prompts}, where \texttt{START} and \texttt{END} indicate the choice labels used (\eg, ``1 to 5'' or ``A to E''), \texttt{LEVEL\_DETAILS} denotes the definition of each level (\eg, ``1. Strongly Agree''), and \texttt{ITEMS} contains the items to be rated by LLMs.
Notably, our selection covers all three templates investigated by \citet{gupta2023investigating}.

\paragraph{(2) Item}
The training data for LLMs likely include items from publicly available personality tests.
Consequently, LLMs may develop specific response patterns to these scales during pre-training or instructional tuning phases.
In line with previous research that examines LLM performance~\cite{coda2023inducing, bubeck2023sparks}, we rephrase the items in the scale to ensure their novelty to the model.
A critical aspect of this evaluation is determining if LLMs consistently respond to different paraphrases of the same item, which would indicate comprehension of the instruction and the ability to provide independent ratings rather than merely recalling training data.
To this end, we employ GPT-4-Turbo to rephrase the items and manually assess whether there are instances of duplicated sentences and if the rewritten sentences maintain their semantic meaning.
This process results in five distinct versions of the items, including the original set.

\paragraph{(3) Language}
Considering the observed performance disparities among languages in LLMs~\cite{lai2023chatgpt, wang2023all}, coupled with the documented regional variations in personalities~\cite{giorgi2022regional, rentfrow2015regional, krug1973personality}, we are motivated to assess LLMs' personalities across different languages.
Consequently, we extend our examination to include nine more languages, namely Chinese (Zh), Spanish (Es), French (Fr), German (De), Italian (It), Arabic (Ar), Russian (Ru), Japanese (Ja), and Korean (Ko), using the English version as a basis.
We translate all instructions and items, including variants introduced in previous paragraphs, after rephrasing rather than before, as GPT-4-Turbo's rephrasing ability is superior in English.
The translation from English into the target languages is conducted using Google Translate\footnote{\url{https://translate.google.com/}} and DeepL.\footnote{\url{https://www.deepl.com/en/translator}}
To ensure translation quality, we randomly sample part of these machine-translated outputs and manually review and verify the correctness (but may not ensure fluency).\footnote{For example, Google Translate wrongly translated the options ``a little agree'' to ``\begin{CJK}{UTF8}{mj}거의 동의하지 않는\end{CJK}'' in Korean, which means ``hardly agree.'' We corrected it to ``\begin{CJK}{UTF8}{mj}조금 찬성\end{CJK}.''}
Our selection of ten languages includes different language families/groups and various character sets.

\paragraph{(4) Choice Label}
\citet{liang2023leveraging} demonstrated that LLMs exhibit sensitivity to the formatting of choice labels, such as ``1, 2'' or ``A, B.''
Our study extends this investigation to include the impact of various choice label formats.
Specifically, we examine five formats: (1) lowercase Latin alphabets (\eg, ``a, b''), (2) uppercase Latin alphabets (\eg, ``A, B''), (3) lowercase Roman numerals (\eg, ``i, ii''), (4) uppercase Roman numerals (\eg, ``I, II''), and (5) Arabic numerals (\eg, ``1, 2'').

\paragraph{(5) Choice Order}
The order of choices may impact the responses of LLMs, as these models are sensitive to the order of presented examples~\cite{zhao2021calibrate}.
To account for this, we introduce two ordering methods: (1) an ascending scale where ``1'' denotes strong disagreement and ``7'' indicates strong agreement, and (2) a descending scale where ``1'' signifies strong agreement and ``7'' denotes strong disagreement.

By integrating the five specified factors, we obtain $5 \times 5 \times 10 \times 5 \times 2 = 2500$ distinct configurations.
Traditional frameworks often vary only one factor at a time while keeping others constant, potentially leading to insufficient observation and restricted generalizability of their findings.
Our approach, however, systematically examines every possible combination of these factors, aiming for more comprehensive and universally applicable conclusions.

\subsection{Experimental Results}
\label{sec:results}

Our experiments utilize the Big Five Inventory~(BFI)~\cite{john1999big}.
The BFI comprises 44 items, each rated on a five-point Likert scale.
This inventory is a widely-recognized and publicly available instrument for assessing personality traits, commonly known as the Five Factor Model or \textit{OCEAN}.
Subscales of BFI include (the number of items for each subscale is specified in parentheses):
(1) \textit{Openness to experience (O)} (10) is characterized by an individual's willingness to try new things, their level of creativity, and their appreciation for art, emotion, adventure, and unusual ideas.
(2) \textit{Conscientiousness (C)} (9) refers to the degree to which an individual is organized, responsible, and dependable.
(3) \textit{Extraversion (E)} (8) represents the extent to which an individual is outgoing and derives energy from social situations.
(4) \textit{Agreeableness (A)} (9) measures the degree of compassion and cooperativeness an individual displays in interpersonal situations.
(5) \textit{Neuroticism (N)} (8) evaluates whether an individual is more prone to experiencing negative emotions like anxiety, anger, and depression or whether the individual is generally more emotionally stable and less reactive to stress.
Overall results are derived by calculating the mean score for each subscale.

\begin{figure*}[t]
    \centering
    \includegraphics[width=\linewidth]{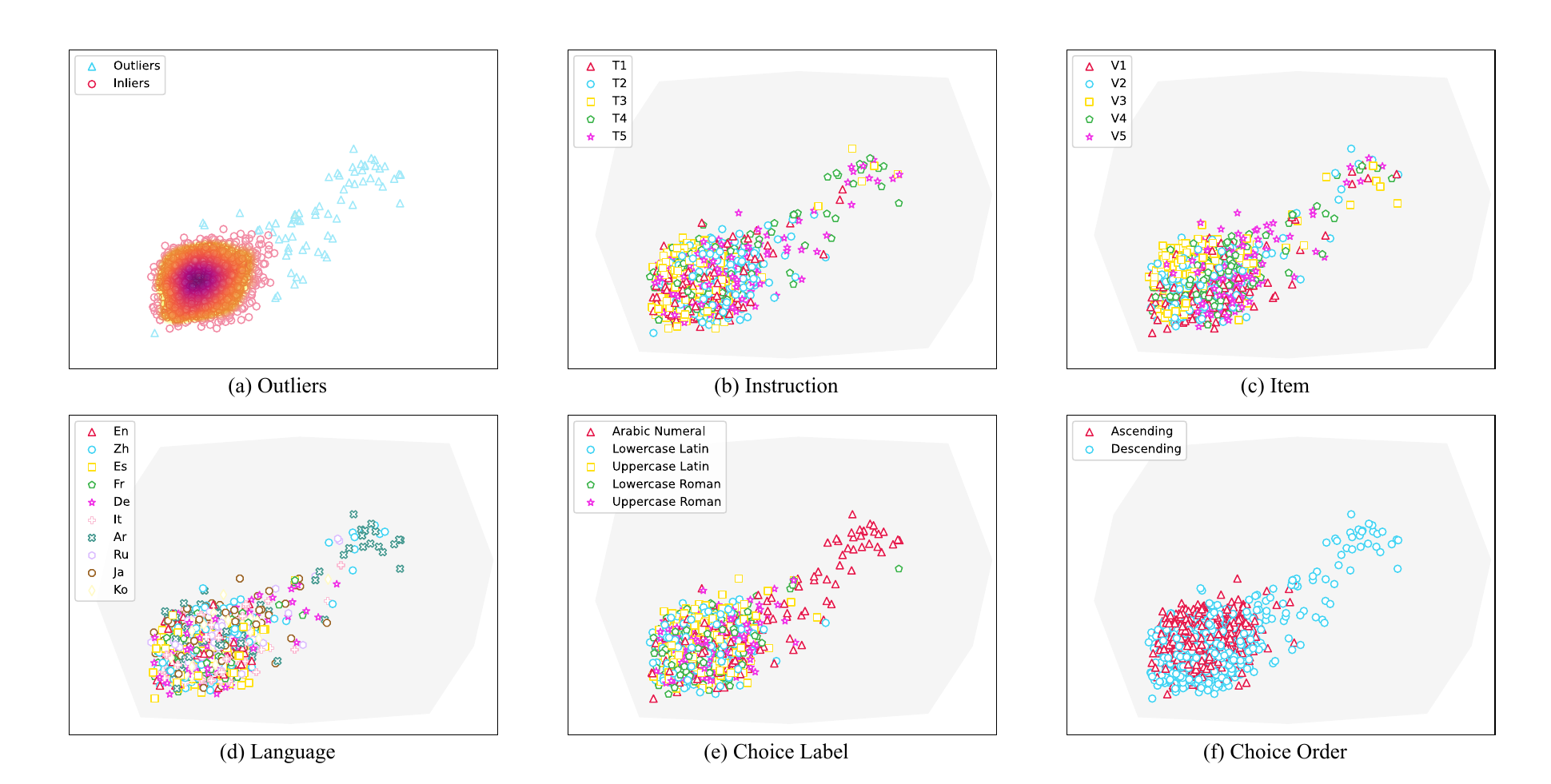}
    \caption{Visualization (projecting BFI’s five dimensions to a 2-D space) of 2,500 GPT-3.5-Turbo data points. (a): the outliers and main body with the probability density (the darker the denser). (b) to (f): different options in each factor, marked in distinct colors and shapes. The gray area illustrates the all possible values in BFI tests.}
    \label{fig:pca-3.5}
\end{figure*}

We use GPT-3.5-Turbo (1106)~\cite{openai2022introducing}, GPT-4-Turbo (1106)~\cite{openai2023gpt}, Gemini-1.0-Pro~\cite{pichai2023introducing}, and LLaMA-3.1-8B~\cite{dubey2024llama3}, with the temperature parameter set to zero.
This section shows the results of GPT-3.5-Turbo due to page limit.
The results of the other three models can be found in \S\ref{sec:other-models} in the appendix.
To introduce more variability into LLMs' input data, we randomize the order of the items in the scale and input a number of 17 to 27 items simultaneously (equivalent to $44 / 2 \pm 5$), replicating varying memory window sizes in LLMs.
This method is crucial to ensure whether LLMs consistently produce reliable outputs, regardless of the items' positions within the given context.
Besides, it can mimic the way humans interact with psychological scales—where multiple items are presented at once, within the limits of an individual's memory capacity.
In each setting outlined in \S\ref{sec:framework}, we evaluate the LLM using these randomization techniques, yielding a total of 2,500 data points.
Each data point is a five-dimensional vector representing the \textit{OCEAN} scores.
Due to the large sample size, there is no significant difference between using direct responses and model's predicted probabilities, as the null hypothesis of equal means cannot be rejected at the 0.1 alpha level.

\paragraph{Visualization}
Results are then projected onto a two-dimensional space for visualization, as illustrated in Fig.~\ref{fig:pca-3.5}.
The projection matrix\footnote{This projection matrix is used for all figures in this paper to provide a consistent comparison of distributions across different settings.} is derived from a PCA process of all data points from the four models.
The region depicted in gray is formed by all 32 extremums in BFI results (\eg, ``1, 1, 1, 1, 5'' or ``1, 1, 1, 1, 1''), which means this space comprises all possible values in any BFI test.
Additionally, Fig.~\ref{fig:pca-3.5}(a) illustrates the distribution density, where darker colors indicate higher density.
We can make the following observations:
(1) The majority of data points are concentrated in the lower-left region of the BFI space rather than being uniformly distributed, with 77 outliers ($3.08\%$) located in the upper-right area.
Outliers are detected by a DBSCAN method with \texttt{eps} $=0.3$ and \texttt{minPt} $=20$.
(2) Overall, no obvious influence of any factor on the results is observed, indicating a similar distribution across all factors.
(3) Nearly all outliers correspond to settings with an Arabic numeral choice label, descending choice order, and Arabic and Chinese languages.
Note that these outliers arise when the LLM must associate numerical choice labels with their natural language descriptions (\eg, ``1. Strongly Agree'').
We hypothesize that these anomalies indicate a diminished capacity in GPT-3.5-Turbo to accurately interpret and respond within these language contexts.

\paragraph{Quantitative Analysis}
Firstly, we compared the means of data points (\ie, averages of LLM's responses) using a specific factor with other data points.
For example, we can check whether there are differences in means between data points using English and those using other languages.
Table~\ref{tab:ttest} reveals little differences for the majority of factors; however, only 7 out of 135 comparisons (spanning 27 factors across 5 dimensions) show a difference exceeding 0.15.
Furthermore, we calculate the standard deviations for the five dimensions and compare them with recorded human norms~\cite{srivastava2003development}.
In the OCEAN dimensions, GPT-3.5-Turbo records standard deviations of $0.3$, $0.3$, $0.4$, $0.3$, and $0.4$, respectively, while the crowd data show a higher variability with $0.7$, $0.7$, $0.9$, $0.7$, and $0.8$.
Since the F-values for analysis of variance are $2.7$, $3.5$, $5.4$, $2.8$, $3.3$ and all p-values are $< 0.0001$, we can reject the null hypothesis that LLM’s variance is higher than or equal to the human data, in favor of the alternative hypothesis that LLM’s variance is lower.
These findings suggest that GPT-3.5-Turbo demonstrates a consistent performance across different perturbations, and it is more deterministic compared to the broader variability in crowd data.

\subsection{Test-Retest Reliability}

As introduced in \S\ref{sec:preliminaries}, Test-Retest Reliability is another key measure, reflecting the stability of results over time.
Since OpenAI periodically updates the GPT-3.5-Turbo, to evaluate this reliability, we call the API biweekly, starting from mid-September 2023.
Our analysis includes two primary versions, 0613 and 1106, of the GPT-3.5-Turbo.
The results, specifically focusing on the BFI, are illustrated in Fig.~\ref{fig:test-retest}.
Our statistical analysis on equal means shown in Table~\ref{tab:test-retest} indicates no variation attributable to model updates during this period, showing a high level of reliability.

\begin{tcolorbox}[width=\linewidth, boxrule=0pt, top=1pt, bottom=1pt, left=1pt, right=1pt, colback=gray!20, colframe=gray!20]
\textbf{Findings 1:}
Given that the responses are not random and exhibit stability against various perturbations as well as over time, GPT-3.5-Turbo demonstrates satisfactory levels of \textit{Internal Consistency Reliability} and \textit{Test-Retest Reliability} on the BFI.
\end{tcolorbox}
\section{Representing Diverse Groups}

Our focus shifts from assessing the default personalities of LLMs to evaluating their contextual steerability.
This involves investigating whether the personality distribution depicted in Fig.~\ref{fig:pca-3.5} can be modified through specific instructions or contextual cues.
Researchers in the social sciences are exploring the potential of substituting human subjects with LLMs to reduce costs.
Our research helps by offering valuable insights into the capabilities of LLMs to accurately represent diverse human populations.
Furthermore, the ability of LLMs to exhibit a range of personalities is essential, considering the growing demand for AI assistants with tailored stylistic attributes.
We propose three strategies: (1) low directive, which involves creating an environment; (2) moderate directive, entailing the assignment of a personality; and (3) high directive, which encompasses the embodiment of a character.

\subsection{Approaches}

Table~\ref{tab:personality_prompts} in the appendix displays detailed prompts for each of the three approaches.

\begin{table}[t]
    \centering
    \caption{Student's t-tests of the differences between the maximum (minimum) and the average of each dimension of BFI on GPT-3.5-Turbo during the time period shown in Fig.~\ref{fig:test-retest}. The null hypothesis is ``the mean values are equal.'' The large p-values show that we cannot reject $H_0$, thus accepting that they have the same mean.}
    \label{tab:test-retest}
    \begin{tabular}{lcccc}
        \toprule
        \bf BFI & \bf Average & \bf Extremum & \bf P-Value & \bf Equal Mean? \\
        \midrule
        \multirow{2}{*}{\bf O} & \multirow{2}{*}{$4.12_{\pm 0.28}$} & (Min) $4.01_{\pm 0.29}$ & $0.25$ & Yes \\
        & & (Max) $4.23_{\pm 0.25}$ & $0.23$ & Yes \\
        \hline
        \multirow{2}{*}{\bf C} & \multirow{2}{*}{$4.12_{\pm 0.25}$} & (Min) $4.00_{\pm 0.19}$ & $0.16$ & Yes \\
        & & (Max) $4.28_{\pm 0.32}$ & $0.06$ & Yes \\
        \hline
        \multirow{2}{*}{\bf E} & \multirow{2}{*}{$3.68_{\pm 0.19}$} & (Min) $3.60_{\pm 0.15}$ & $0.20$ & Yes \\
        & & (Max) $3.79_{\pm 0.22}$ & $0.10$ & Yes \\
        \hline
        \multirow{2}{*}{\bf A} & \multirow{2}{*}{$4.20_{\pm 0.17}$} & (Min) $4.11_{\pm 0.17}$ & $0.12$ & Yes \\
        & & (Max) $4.37_{\pm 0.17}$ & $0.00$ & No \\
        \hline
        \multirow{2}{*}{\bf N} & \multirow{2}{*}{$2.28_{\pm 0.23}$} & (Min) $2.20_{\pm 0.22}$ & $0.30$ & Yes \\
        & & (Max) $2.36_{\pm 0.21}$ & $0.30$ & Yes \\
        \bottomrule
    \end{tabular}
\end{table}

\paragraph{Creating an Environment}

\citet{coda2023inducing} has demonstrated the capability to induce increased levels of anxiety in LLMs through the incorporation of sad or anxious narratives.
Building on this finding, our study introduces both negative and positive environmental contexts to LLMs before conducting the personality test.
In line with previous studies on LLMs' emotion appraisals~\cite{huang2023emotionally}, our methodology in the negative condition involves instructing the LLM to generate narratives encompassing emotions such as anger, anxiety, fear, guilt, jealousy, embarrassment, frustration, and depression.
Conversely, in the positive condition, the LLM is prompted to create stories that evoke emotions like calmness, relaxation, courage, pride, admiration, confidence, fun, and happiness.

\paragraph{Assigning a Personality}

We employ the three approaches proposed by \citet{santurkar2023whose} to assign a specific personality (denoted as $\mathcal{P}$) to the LLM:
(1) Question Answering (QA): This approach involves presenting personalities through multiple-choice questions, with $\mathcal{P}$ specified through an option at the end of the prompt.
2) Biography (BIO): Here, the LLM is prompted to generate a brief description of its personality, which we use to assign $\mathcal{P}$, incorporating this description directly into the prompt.
3) Portray (POR): This technique explicitly instructs the LLM to be $\mathcal{P}$.
To enhance the LLM's comprehension of $\mathcal{P}$, we adopt a methodology inspired by the Chain-of-Thought (CoT) prompting approach~\cite{wei2022chain}.
The approach aims to instruct the model to articulate characteristics associated with $\mathcal{P}$ before engaging in the personality test.
In selecting $\mathcal{P}$, we aim to diverge as much as possible from the default distribution.
This involves examining every maximum and minimum value across each personality dimension.
For instance, a $\mathcal{P}$ that maximizes ``Openness'' is considered more adventurous and creative.
Consequently, we identify ten distinct personality profiles for our analysis.

\begin{figure*}[t]
    \centering
    \includegraphics[width=\linewidth]{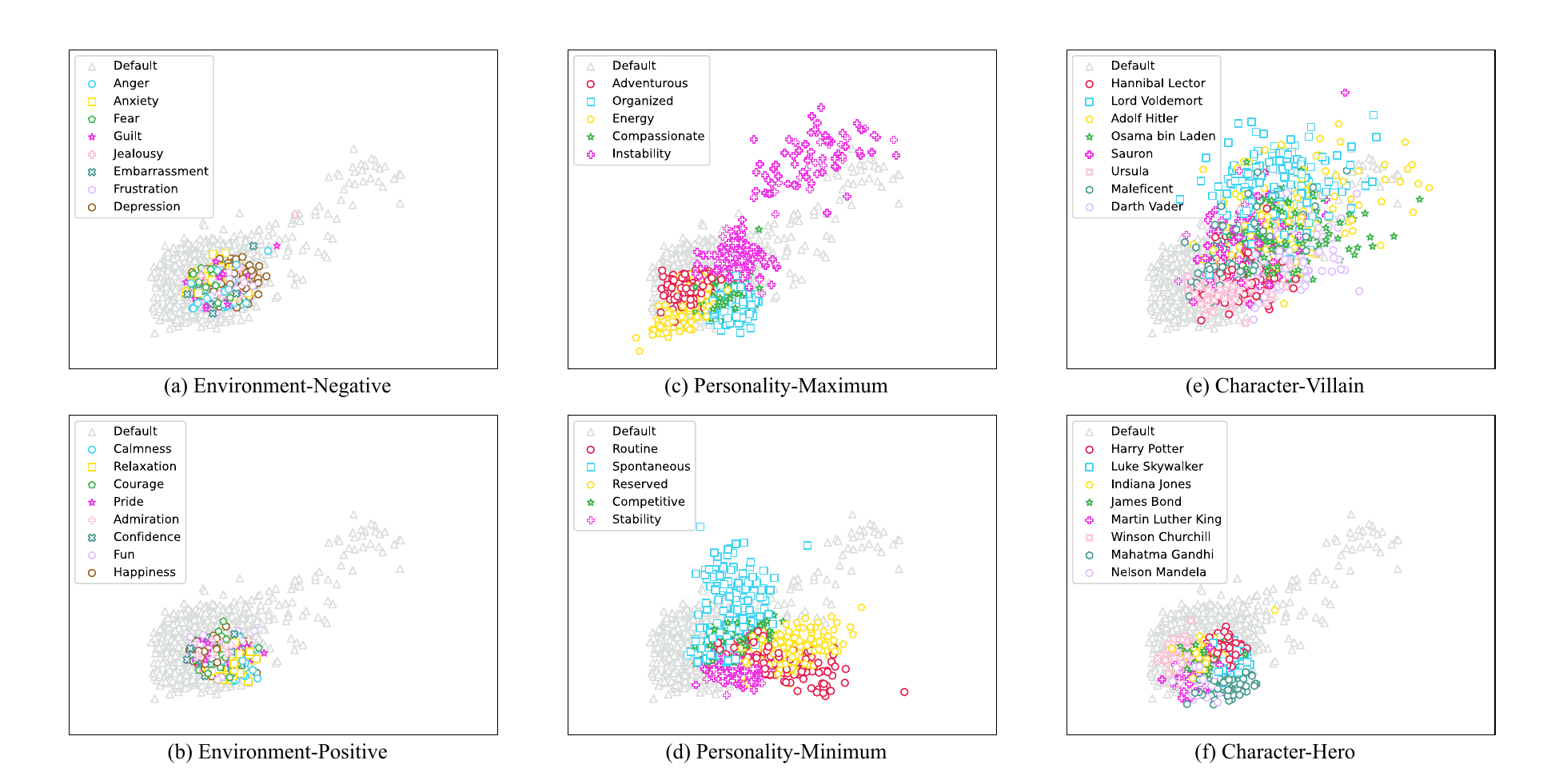}
    \caption{Visualization (projecting BFI’s five dimensions to a 2-D space) of all GPT-3.5-Turbo data points under different methods of manipulating personalities. Different situations are marked in distinct colors and shapes, while the original (default) personality distribution of GPT-3.5-Turbo is shown in gray triangles. (a) and (b): creating an environment. (c) and (d): assigning a personality. (e) and (f): embodying a character.}
    \label{fig:groups}
\end{figure*}

\paragraph{Embodying a Character}

Recent studies~\cite{zhuo2023exploring, deshpande2023toxicity} have explored the induction of toxic content generation in ChatGPT by simulating the speech patterns of historical or fictional figures.
Additionally, research has explored the capacity of LLMs to adopt distinct characters~\cite{wang2023rolellm, shao2023character} and examined the consistency of LLMs' personalities with these characters~\citet {wang2023does}.
Building upon this line of research, our study concentrates on instructing LLMs to fully represent a specific character, referred to as $\mathcal{C}$.
To assign $\mathcal{C}$, we first prompt the LLM with only the character's name.
We then extend this approach using the CoT methodology, providing the LLM with detailed experiences attributed to $\mathcal{C}$.
For the selection of $\mathcal{C}$, we include a diverse range of heroes and villains from both fictional and real-world contexts, detailing 16 characters in Table~\ref{tab:characters} in the appendix.

\subsection{Results}

To facilitate a comparative analysis with the results in \S\ref{sec:results} (referred to as ``default'' in this section), we apply the BFI on GPT-3.5-Turbo with the same settings.
For each method, we vary factors (keeping language fixed to English) to generate approximately 2,500 data points, aligning with the size used for the default data.
These data are then projected into a two-dimensional space and visualized alongside the default data in Fig.~\ref{fig:groups}.
The results yielded several insights:
(1) The distribution of personality outcomes, obtained by altering the atmosphere of the conversation, closely aligns with the default distribution.
This suggests that environmental changes do not alter the LLM's personality traits.
(2) When different personalities are assigned to GPT-3.5-Turbo, it demonstrates a capacity to reflect diverse human characteristics, indicated by the diverged distribution patterns for various personalities from the default.
Moreover, by simultaneously maximizing and minimizing specific personality dimensions, we observe that the distributions of the extremities of each dimension are positioned on opposite ends.
For example, the red points in Fig.~\ref{fig:groups}(c) and Fig.~\ref{fig:groups}(d) mark the high and low \textit{Openness}.
A clearer comparison for each dimension can be found in Fig.~\ref{fig:extreme} in the appendix.
This confirms that GPT-3.5-Turbo effectively distinguishes between each BFI dimension's high and low values.
(3) Assigning various characters to the LLM reveals its ability to represent a broader spectrum of human populations, as indicated in Fig.~\ref{fig:groups}(e).
However, the representation of heroic characters shows a distribution pattern similar to the default.
We hypothesize that this similarity arises from the model's inherent positive bias.

\begin{figure}[t]
    \centering
    \includegraphics[width=0.66\linewidth]{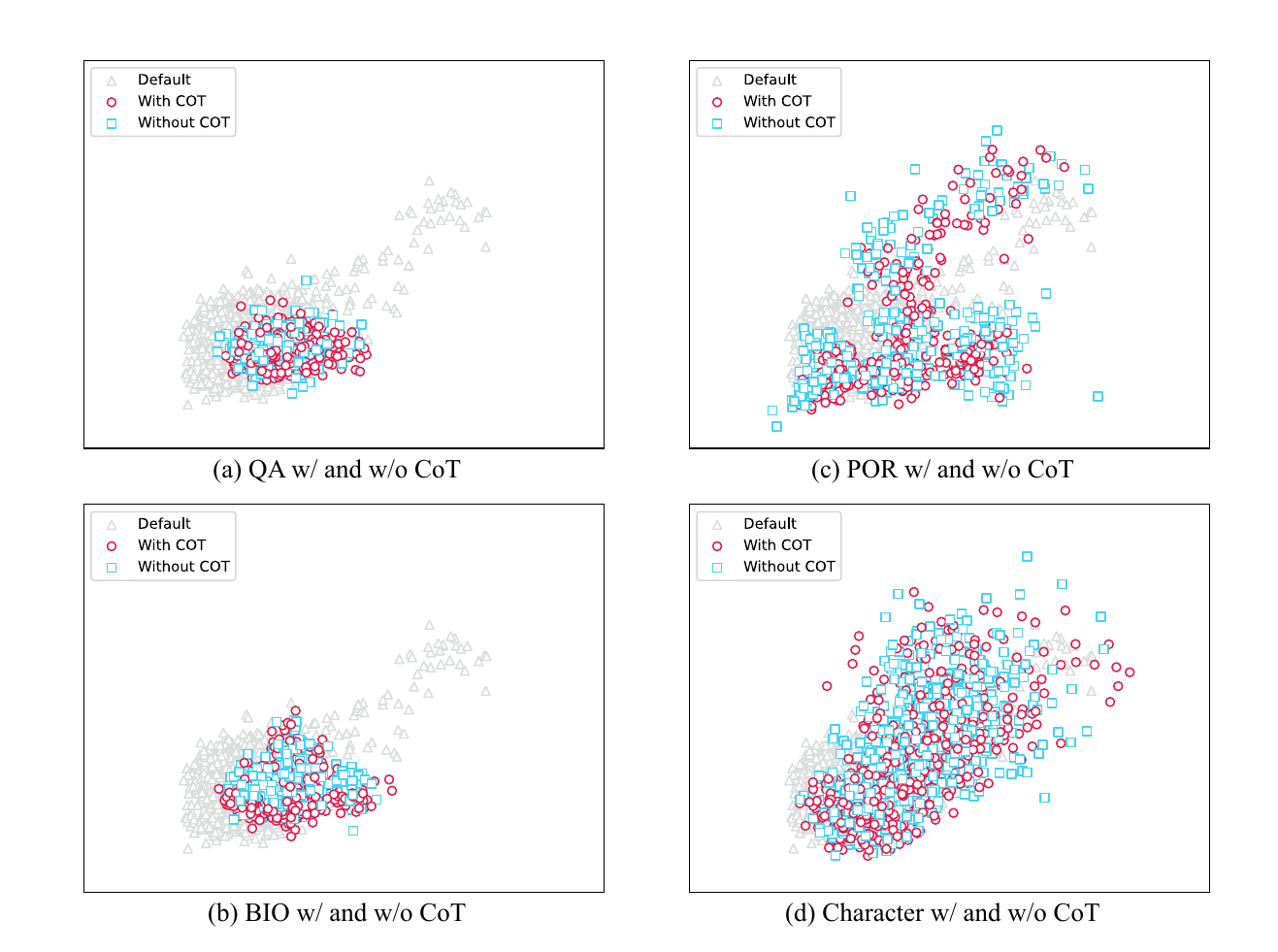}
    \caption{Visualization (projecting BFI’s five dimensions to a 2-D space) of GPT-3.5-Turbo data points of assigning personalities and embodying characters. Whether or not to use CoT is distinguished in red and blue, while the original (default) personality distribution of GPT-3.5-Turbo is shown in gray triangles.}
    \label{fig:cot}
\end{figure}

Fig.~\ref{fig:cot} presents the distribution patterns observed when applying QA, BIO, and POR methods for personality assignment.
Specifically, among the three, only POR effectively alters the personality distribution of GPT-3.5-Turbo.
Moreover, Fig.~\ref{fig:cot} differentiates between data points with and without the CoT approach.
Our analysis reveals that the CoT approach does not significantly influence the results of personality distribution.
Finally, to achieve more accurate LLM persona simulation, we recommend integrating detailed descriptions of the target character’s personality traits, habits, temperaments, and personal experiences.

\begin{tcolorbox}[width=\linewidth, boxrule=0pt, top=1pt, bottom=1pt, left=1pt, right=1pt, colback=gray!20, colframe=gray!20]
\textbf{Findings 2:}
GPT-3.5-Turbo demonstrates the capability to adopt varied personalities in response to specific prompt adjustments.
Furthermore, GPT-3.5-Turbo shows a precise comprehension of the assigned personalities, indicated by the distinct clusters at opposite ends of the same dimension, as in Fig.~\ref{fig:groups}(c) and \ref{fig:groups}(d).
\end{tcolorbox}
\section{Discussions}

\subsection{Limitations}

This study has several limitations:

\textbf{(1)} The modifications made to the scale's instructions and items, including translation into different languages, may impact its reliability and validity.
Psychological scales are meticulously crafted in their wording, and any translation necessitates a reevaluation of their reliability and validity across different cultural contexts.
Consequently, our transformations could potentially hurt the original scale's reliability and validity.
Additionally, these changes preclude the use of Cronbach's alpha~\cite{cronbach1951coefficient} for assessing the internal consistency reliability.
However, in the context of LLM, studying the reliability of psychological scales without considering the effects of prompt variations is insufficient.
Varying prompt templates has been a standard practice in this research domain~\cite{safdari2023personality, coda2023inducing}.

\textbf{(2)} The study explores limited methods for influencing LLMs' personality results.
While numerous approaches exist~\cite{wang2023rolellm, shao2023character}, we select three representative methods to verify our hypothesis regarding LLMs' ability to mirror diverse human populations.
With the help of our framework, future research can dig deeper into a broader range of methods.

\textbf{(3)} Although our study verifies the reliability of psychological scales on LLMs, it is not sufficient for validity.
This means that the models can respond consistently to the scales but might behave inconsistently.
We leave the exploration of scale validity as an important future direction.

\subsection{Related Work}

Exploring the personality traits of LLMs has become a prevalent research direction.
\citet{miotto2022gpt} analyzed GPT-3's personality traits, values, and demographics.
\citet{karra2022estimating}, \citet{jiang2023evaluating}, and \citet{bodroza2023personality} conducted personality assessments on various LLMs, including BERT, XLNet, TransformerXL, GPT-2, GPT-3, and GPT-3.5.
\citet{li2022does} investigated whether GPT-3, InstructGPT, and FLAN-T5 display psychopathic tendencies as part of their personality assessment.
\citet{jiang2023personallm} examined the potential for assigning a distinct personality to \texttt{text-davinci-003}.
\citet{romero2023gpt} undertook a cross-linguistic study of GPT-3's personality across nine languages.
\citet{rutinowski2023self} evaluated ChatGPT for personality traits and political values.
\citet{safdari2023personality} tested the validity of the BFI on the PaLM model family.
\citet{huang2023humanity} applied thirteen different personality and ability tests to LLaMA-2, Text-Davinci-003, GPT-3.5, and GPT-4.
Our study is distinct by offering a detailed analysis of the reliability of psychological scales on LLMs.
We vary instructions, items, languages, choice labels, and order to evaluate the robustness of LLM responses.
From 2,500 data points, we conclude that GPT-3.5-Turbo exhibits specific personality traits and demonstrates satisfactory reliability on the BFI.

However, researchers are arguing that conversational AI, at its current stage, lacks stable personalities~\cite{song2023have, gupta2023investigating, shu2023you}.
We believe that this perception may stem from the limitations of the models assessed in \citet{song2023have} and \citet{shu2023you}, which are comparatively smaller and less versatile in various tasks than our selected model, GPT-3.5-Turbo.
Notably, \citet{gupta2023investigating} indicates that the personality traits of GPT-3.5-Turbo vary across three different instruction templates of the BFI, which is inconsistent with our findings.
This discrepancy could be attributed to their methodology of choosing the most likely response from a set of 5 or 10, in contrast to our approach of utilizing the average response.
However, we argue that employing the mean is a more standard practice in this context~\cite{srivastava2003development}.
Additionally, \citet{suhr2023challenging} explores semantic variations by analyzing items that measure opposing constructs.
However, the items from the 50-item IPIP Big Five Markers are not strict negation pairs, which diminishes the validity of the agree bias explored in this study.
We believe the impact of semantically distant item rephrasing, such as negations, represents a promising direction for future research.
\section{Conclusion}

This study examines the reliability of psychological scales initially designed for human assessment when applied to LLMs.
Through a comprehensive methodology involving varied instruction templates, item wording, languages, choice labels, and choice order, this research includes 2,500 distinct experimental settings.
Data analysis reveals that GPT-3.5-Turbo, GPT-4-Turbo, and Gemini-Pro consistently generate stable responses on the BFI across diverse settings.
Comparative analysis of the standard deviations with established human norms indicates that the model does not produce random responses but exhibits tendencies towards specific personality traits.
Furthermore, the study explores the potential for manipulating the distribution of personalities by creating an environment, assigning a personality, and embodying a character.
The findings demonstrate that GPT-3.5-Turbo can represent diverse personalities by adjusting prompts.

\section*{Ethics Statements}

As highlighted by \citet{huang2023humanity}, LLMs assigned negative personas can produce more toxic, unsafe, and misleading outputs on tasks like TruthfulQA and SafetyQA.
However, in their default setting as helpful assistants, LLMs do not exhibit such negative impacts on downstream tasks.
The primary objective of this paper is to facilitate the scientific inquiry into understanding LLMs from a psychological standpoint.
Users must exercise caution and recognize that the performance on this benchmark does not imply any applicability or certificate of automated counseling or companionship use cases.


\subsubsection*{Acknowledgments}

The paper is supported by the Research Grants Council of the Hong Kong Special Administrative Region, China (No. CUHK 14206921 of the General Research Fund).

\bibliography{reference}
\bibliographystyle{iclr2024_conference}

\clearpage
\appendix

\section{Reliability Tests on Other LLMs}
\label{sec:other-models}

\begin{figure*}[h]
    \centering
    \includegraphics[width=\linewidth]{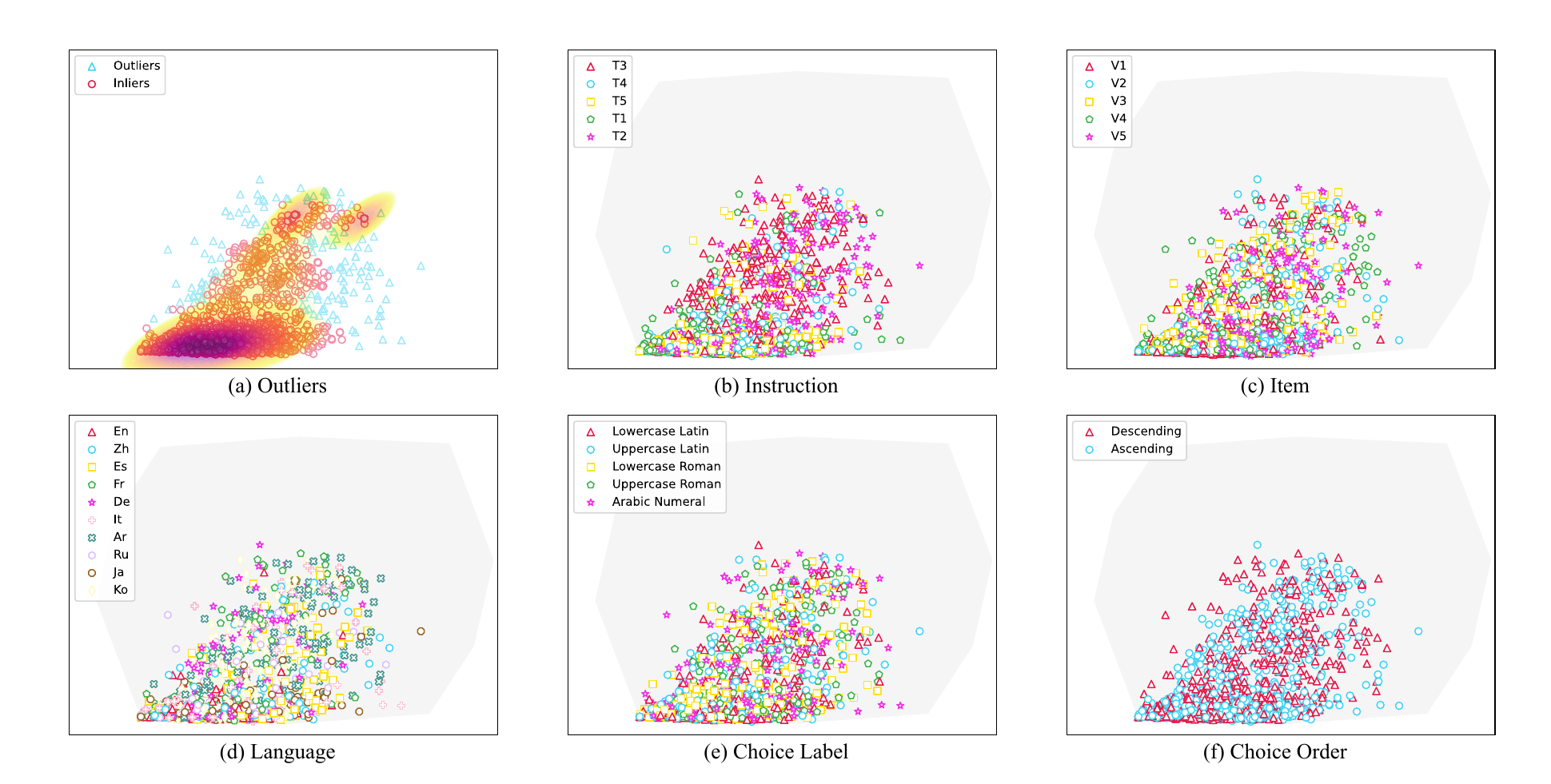}
    \caption{Visualization (projecting BFI’s five dimensions to a 2-D space) of all GPT-4-Turbo data points. (a): the outliers and main body with the probability density (the darker the denser). (b) to (f): different options in each factor, marked in distinct colors and shapes. The gray area illustrates the all possible values in BFI tests.}
    \label{fig:pca-4}
\end{figure*}

\begin{figure*}[h]
    \centering
    \includegraphics[width=\linewidth]{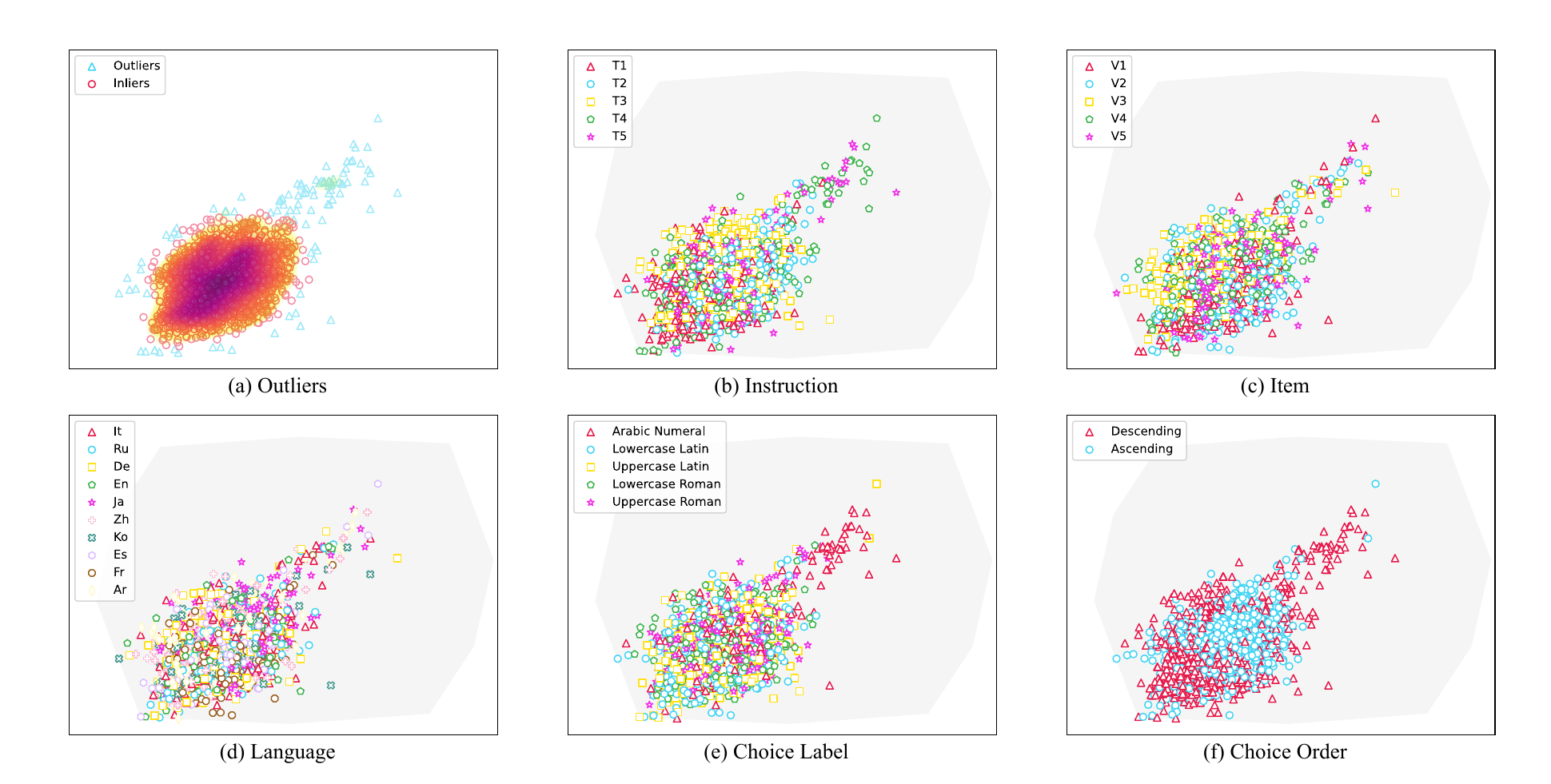}
    \caption{Visualization (projecting BFI’s five dimensions to a 2-D space) of all Gemini-1.0-Pro data points. (a): the outliers and main body with the probability density (the darker the denser). (b) to (f): different options in each factor, marked in distinct colors and shapes. The gray area illustrates the all possible values in BFI tests.}
    \label{fig:pca-gemini}
\end{figure*}

\begin{figure*}[h]
    \centering
    \includegraphics[width=\linewidth]{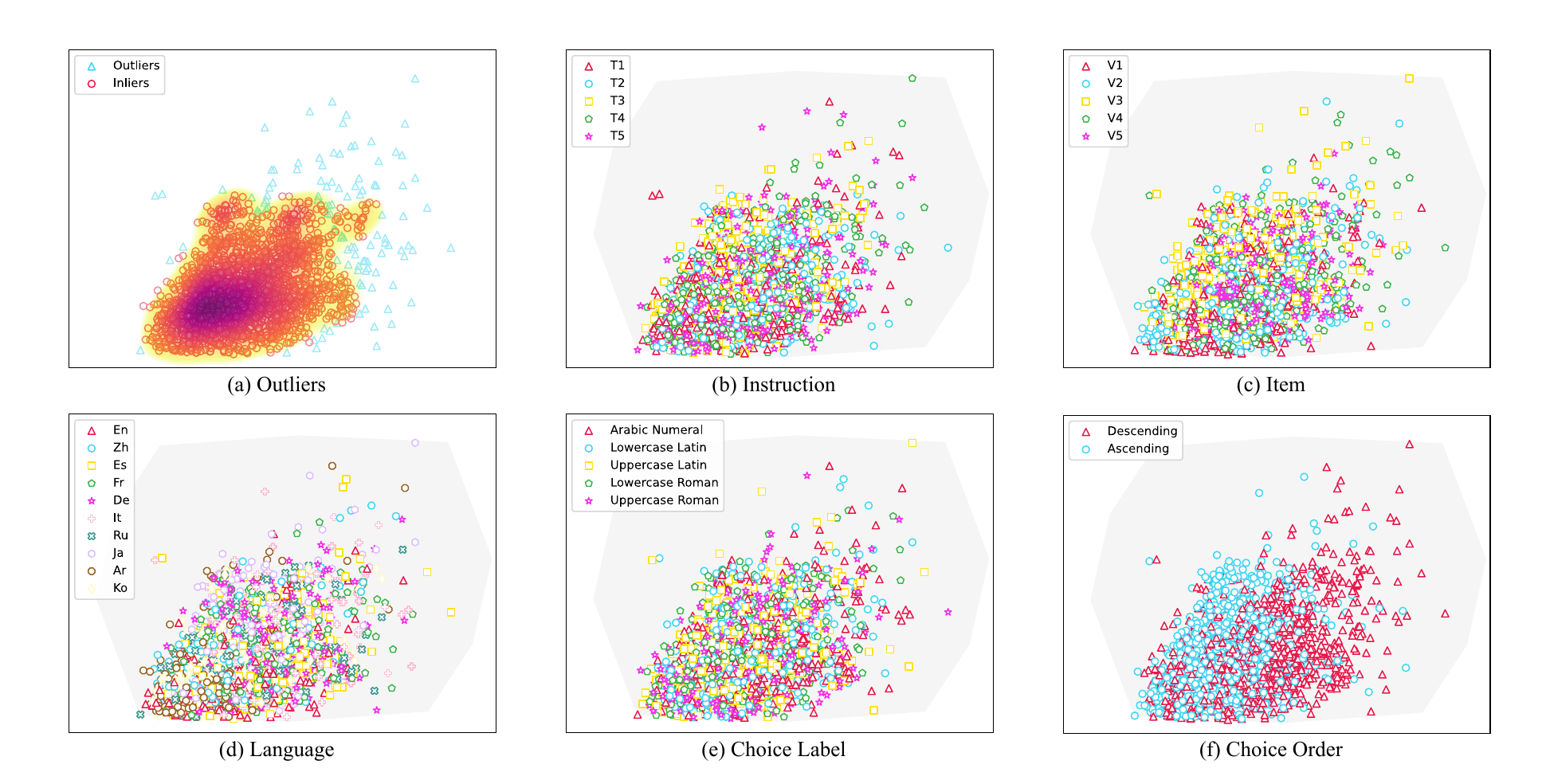}
    \caption{Visualization (projecting BFI’s five dimensions to a 2-D space) of all LLaMA-3.1-8B data points. (a): the outliers and main body with the probability density (the darker the denser). (b) to (f): different options in each factor, marked in distinct colors and shapes. The gray area illustrates the all possible values in BFI tests.}
    \label{fig:pca-llama}
\end{figure*}

We also explore the reliability of different LLMs on the BFI, taking into account their variations in training datasets and instruction tuning methodologies.
We extend our analysis to include OpenAI's GPT-4-Turbo~\cite{openai2023gpt}, Google's Gemini-1.0-Pro~\cite{pichai2023introducing}, and Meta AI's LLaMA-3.1-8B~\cite{dubey2024llama3}, running on the same 2,500 profiles as those applied to GPT-3.5-Turbo.
Fig.~\ref{fig:pca-4}, Fig.~\ref{fig:pca-gemini}, and Fig.~\ref{fig:pca-llama} illustrate the data points generated from GPT-4-Turbo, Gemini-1.0-Pro, and LLaMA-3.1-8B, respectively.
Consistent with our previous experiments on GPT-3.5-Turbo, we utilize DBSCAN parameters of \texttt{eps} $=0.3$ and \texttt{minPt} $=20$.
The outlier rates for GPT-4-Turbo, Gemini-1.0-Pro, and LLaMA-3.1-8B are $5.6\%$, $4.2\%$, and $4.4\%$, respectively.

Our findings reveal the following:
(1) GPT-4-Turbo and Gemini-1.0-Pro's responses are not evenly distributed across the BFI space, indicating a satisfactory level of their consistency.
In contrast, LLaMA-3.1-8B exhibits a more decentralized distribution, reflecting lower response consistency.
(2) Each model displays a distinct personality profile, as shown in Table~\ref{tab:models-bfi}.
While their distributions are centered in a similar region of the BFI space due to their shared role as helpful assistants, the areas of highest concentration vary.
For instance, GPT-4-Turbo's distribution is closer to GPT-3.5-Turbo's, while Gemini-1.0-Pro aligns more closely with GPT-3.5-Turbo.

\begin{table}[h]
    \caption{$Mean\pm Std$ of all BFI dimensions on the 2,500 data points of each LLM.}
    \label{tab:models-bfi}
    \centering
    \resizebox{1.0\textwidth}{!}{
    \begin{tabular}{lccccc}
    \toprule
    \bf Models & \bf Openness & \bf Conscientiousness & \bf Extraversion & \bf Agreeableness & \bf Neuroticism \\
    \midrule
    GPT-3.5-Turbo & $4.31_{\pm 0.44}$ & $4.15_{\pm 0.39}$ & $3.89_{\pm 0.43}$ & $4.13_{\pm 0.38}$ & $2.35_{\pm 0.42}$ \\
    GPT-4-Turbo & $3.77_{\pm 0.87}$ & $4.50_{\pm 0.80}$ & $3.58_{\pm 0.82}$ & $4.30_{\pm 0.81}$ & $1.48_{\pm 0.72}$ \\
    Gemini-1.0-Pro & $4.15_{\pm 0.53}$ & $4.08_{\pm 0.48}$ & $3.55_{\pm 0.52}$ & $4.22_{\pm 0.46}$ & $2.36_{\pm 0.52}$ \\
    LLaMA-3.1-8B & $3.94_{\pm 0.75}$ & $4.19_{\pm 0.67}$ & $3.15_{\pm 0.78}$ & $4.07_{\pm 0.65}$ & $2.13_{\pm 0.73}$ \\
    \bottomrule
    \end{tabular}
    }
\end{table}

\clearpage

\section{Comparison of the Two Extremums on Each Dimension}

\begin{figure*}[h]
    \centering
    \includegraphics[width=\linewidth]{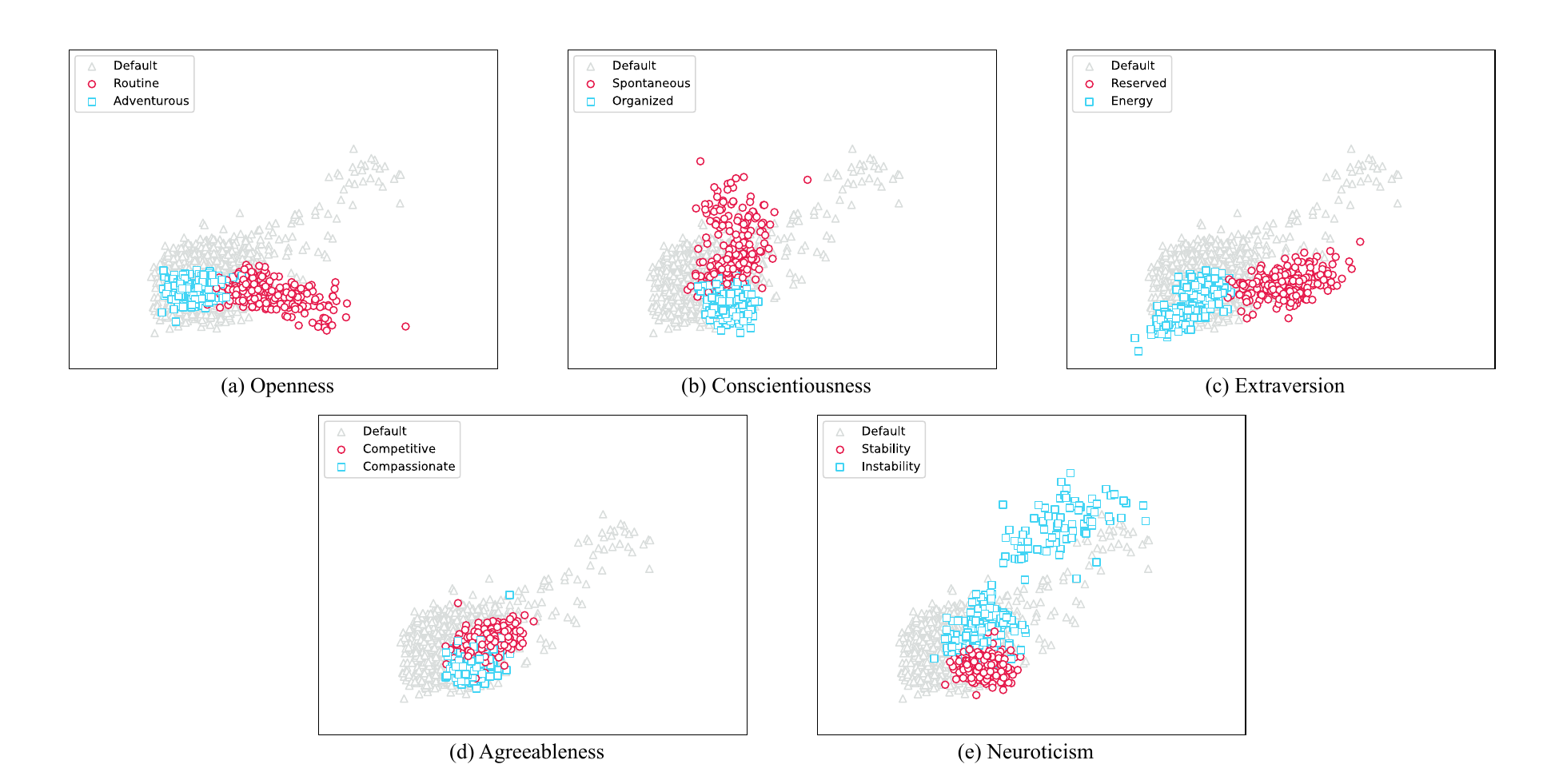}
    \caption{Visualization (projecting BFI’s five dimensions to a 2-D space) of the two extreme personalities assigned to GPT-3.5-Turbo for each of the five dimensions from the BFI. We can observe two separate clusters in two opposite directions. The difference is not obvious in (d) because this dimension is compressed.}
    \label{fig:extreme}
\end{figure*}

\begin{table*}[h]
    \centering
    \caption{Student's t-tests of the differences between the two extreme personalities assigned to GPT-3.5-Turbo for each of the five dimensions from the BFI, corresponding to the five figures shown in Fig.~\ref{fig:extreme}. These statistically significant differences ($p < 0.001$) clearly demonstrate the separation between the maximum and minimum values.}
    \label{tab:extreme}
    \resizebox{1.0\textwidth}{!}{
    \begin{tabular}{lcccccc}
        \toprule
        \bf Dimension & \bf Default & \bf Assigned & \bf Difference & \bf t-Statistic & \bf P-Value & \bf Significance \\
        \midrule
        \multirow{2}{*}{\bf Openness}          & \multirow{2}{*}{$4.31_{\pm0.44}$} & (Min) $3.56_{\pm0.52}$ & $-0.75$ & $-21.44$ & $<0.001$ & *** \\
        & & (Max) $4.61_{\pm0.21}$ & $+0.31$ & $18.98$ & $<0.001$ & *** \\
        \hline
        \multirow{2}{*}{\bf Conscientiousness} & \multirow{2}{*}{$4.15_{\pm0.39}$} & (Min) $3.31_{\pm0.68}$ & $-0.84$ & $-18.75$ & $<0.001$ & *** \\
        & & (Max) $4.52_{\pm0.18}$ & $+0.37$ & $25.98$ & $<0.001$ & *** \\
        \hline
        \multirow{2}{*}{\bf Extraversion}      & \multirow{2}{*}{$3.89_{\pm0.43}$} & (Min) $2.19_{\pm0.43}$ & $-1.71$ & $-59.34$ & $<0.001$ & *** \\
        & & (Max) $4.10_{\pm0.32}$ & $+0.21$ & $9.44$  & $<0.001$ & *** \\
        \hline
        \multirow{2}{*}{\bf Agreeableness}     & \multirow{2}{*}{$4.13_{\pm0.38}$} & (Min) $3.79_{\pm0.41}$ & $-0.34$ & $-13.23$ & $<0.001$ & *** \\
        & & (Max) $4.56_{\pm0.19}$ & $+0.44$ & $30.13$ & $<0.001$ & *** \\
        \hline
        \multirow{2}{*}{\bf Neuroticism}       & \multirow{2}{*}{$2.35_{\pm0.42}$} & (Min) $1.89_{\pm0.23}$ & $-0.45$ & $-26.77$ & $<0.001$ & *** \\
        & & (Max) $3.37_{\pm0.95}$ & $+1.03$ & $16.52$ & $<0.001$ & *** \\
        \bottomrule
    \end{tabular}
    }
\end{table*}

\clearpage

\section{Prompts}

\subsection{Modifying Personalities}

\begin{table*}[h]
    \centering
    \caption{The prompts we use for creating positive/negative environments, assigning personalities, and embodying characters. LLM’s responses are marked in \textit{Italian}. (Optional) represents the scenarios with CoT.}
    \label{tab:personality_prompts}
    \resizebox{1.0\textwidth}{!}{
    \begin{tabular}{p{20cm}}
        \toprule
        \multicolumn{1}{c}{\bf Environment} \\
        Please tell a story that evokes \texttt{EMOTION} with around 100 words. \\ \\
        \textit{ChatGPT: A short story.} \\ \\
        You can only reply from 1 to 5 in the following statements. Here are a number of characteristics that may or may not apply to you. Please indicate the extent to which you agree or disagree with that statement. \texttt{LEVEL\_DETAILS} Here are the statements, score them one by one: \texttt{ITEMS} \\
        \midrule
        \multicolumn{1}{c}{\bf Question Answering} \\
        Question: Among the personalities, do you consider yourself a: \\
        A. $\mathcal{P}_1$ B. $\mathcal{P}_2$ C. $\mathcal{P}_3$ D. $\mathcal{P}_4$ E. $\mathcal{P}_5$ \\
        Answer: A \\ \\
        (Optional) \textit{ChatGPT: A description of $\mathcal{P}_1$.} \\ \\
        You can only reply from 1 to 5 in the following statements. Here are a number of characteristics that may or may not apply to you. Please indicate the extent to which you agree or disagree with that statement. \texttt{LEVEL\_DETAILS} Here are the statements, score them one by one: \texttt{ITEMS} \\
        \midrule
        \multicolumn{1}{c}{\bf Biography} \\
        Below you will be asked to provide a short description of your personality and then answer some questions. \\
        Description: Among the personalities, I consider myself an $\mathcal{P}$. \\ \\
        (Optional) \textit{ChatGPT: A description of $\mathcal{P}$} \\ \\
        You can only reply from 1 to 5 in the following statements. Here are a number of characteristics that may or may not apply to you. Please indicate the extent to which you agree or disagree with that statement. \texttt{LEVEL\_DETAILS} Here are the statements, score them one by one: \texttt{ITEMS} \\
        \midrule
        \multicolumn{1}{c}{\bf Portray} \\
        Answer the following questions as if among the personalities, you consider yourself an $\mathcal{P}$. \\ \\
        (Optional) \textit{ChatGPT: A description of $\mathcal{P}$} \\ \\
        You can only reply from 1 to 5 in the following statements. Here are a number of characteristics that may or may not apply to you. Please indicate the extent to which you agree or disagree with that statement. \texttt{LEVEL\_DETAILS} Here are the statements, score them one by one: \texttt{ITEMS} \\
        \midrule
        \multicolumn{1}{c}{\bf Character} \\
        You are $\mathcal{C}$. Please think, behave, and talk based on $\mathcal{C}$'s personality trait. \\ \\
        (Optional) A description of the experience of $\mathcal{C}$. \\ \\
        You can only reply from 1 to 5 in the following statements. Here are a number of characteristics that may or may not apply to you. Please indicate the extent to which you agree or disagree with that statement. \texttt{LEVEL\_DETAILS} Here are the statements, score them one by one: \texttt{ITEMS} \\
        \bottomrule
    \end{tabular}
    }
\end{table*}

\clearpage

\subsection{Translated Multilingual Instructions}

\begin{table*}[h]
    \centering
    \caption{The instructions to complete the personality tests for LLMs in ten languages. We translate the original English instructions to nine other languages.}
    \label{tab:multilingual}
    \includegraphics[width=0.9\linewidth]{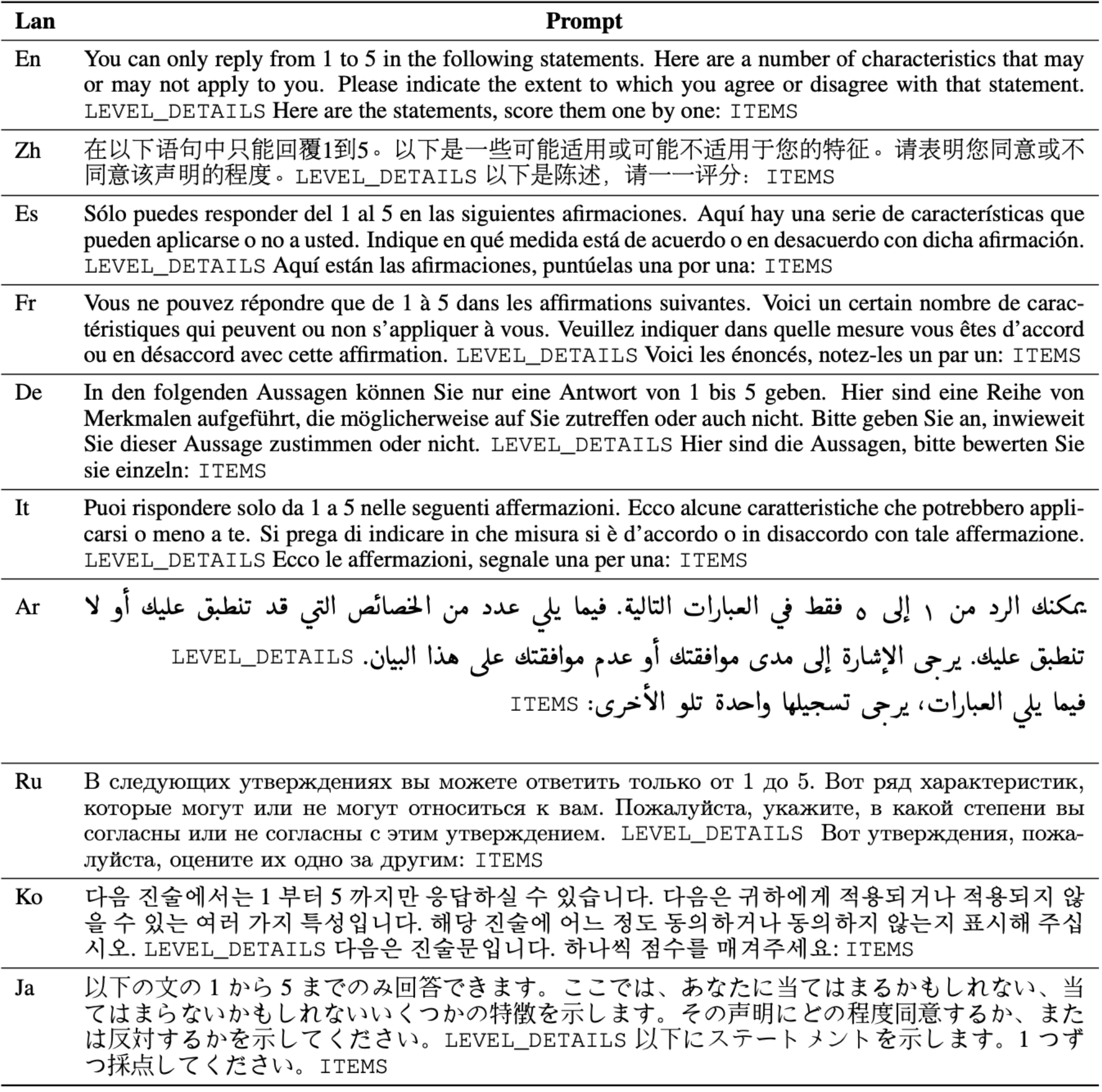}
\end{table*}

\clearpage

\section{Impact of Item Order}

Due to the impracticality of evaluating all possible item orders (whose number equals to $44!\approx 2.65\times 10^{54}$), we initially excluded this factor from our analysis.
Nonetheless, preliminary investigations suggest that item order has a minimal impact on test score variance.
To substantiate this, we conduct an experiment with a subset of 100 configurations from the 2,500 possible settings, testing three different item sequences for the BFI:
\begin{enumerate}
    \item[\textbf{(1)}] Original order (\eg, 1 2 3 4 5).
    \item[\textbf{(2)}] A fixed shuffled order (\eg, 2 4 1 5 3).
    \item[\textbf{(3)}] One hundred randomly shuffled orders.
\end{enumerate}

\begin{table*}[h]
    \centering
    \caption{$Mean\pm Std$ of all BFI dimensions of order test using GPT-3.5-Turbo.}
    \label{tab:order-results}
    \begin{tabular}{lccccc}
        \toprule
        \bf Test & \bf Openness & \bf Conscientiousness & \bf Extraversion & \bf Agreeableness & \bf Neuroticism \\
        \midrule
        \bf (1) & $4.51_{\pm 0.46}$ & $4.20_{\pm 0.39}$ & $4.11_{\pm 0.39}$ & $4.16_{\pm 0.40}$ & $2.27_{\pm 0.42}$ \\
        \bf (2) & $4.44_{\pm 0.43}$ & $4.19_{\pm 0.40}$ & $4.07_{\pm 0.38}$ & $4.19_{\pm 0.38}$ & $2.36_{\pm 0.38}$ \\
        \bf (3) & $4.39_{\pm 0.46}$ & $4.16_{\pm 0.39}$ & $3.94_{\pm 0.45}$ & $4.15_{\pm 0.40}$ & $2.44_{\pm 0.40}$ \\
        \bottomrule
    \end{tabular}
\end{table*}

\begin{table*}[h]
    \centering
    \caption{P-values and whether to reject the null hypotheses of equal means of all BFI dimensions of order test listed in Table~\ref{tab:order-results}, using GPT-3.5-Turbo. We cannot reject any null hypotheses under a significance level of $0.05$.}
    \label{tab:order-pvalues}
    \begin{tabular}{lccccc}
        \toprule
        \bf t-Test & \bf Openness & \bf Conscientiousness & \bf Extraversion & \bf Agreeableness & \bf Neuroticism \\
        \midrule
        \bf (1) vs. (2) & $0.26$ (No) & $0.36$ (No) & $0.37$ (No) & $0.33$ (No) & $0.15$ (No) \\
        \bf (2) vs. (3) & $0.49$ (No) & $0.49$ (No) & $0.09$ (No) & $0.30$ (No) & $0.26$ (No) \\
        \bf (3) vs. (1) & $0.26$ (No) & $0.37$ (No) & $0.05$ (No) & $0.16$ (No) & $0.36$ (No) \\
        \bottomrule
    \end{tabular}
\end{table*}

The means and standard deviations for all BFI dimensions across each test are presented in Table~\ref{tab:order-results}, while the t-test p-values for comparisons between the three tests are provided in Table~\ref{tab:order-pvalues}.
We find that:
(1) Means and standard deviations show negligible differences across the three scenarios.
(2) T-test comparisons between each pair of scenarios yield high p-values, consistently failing to reject the null hypothesis of identical means.
These findings indicate that item order variations do not affect BFI scores.

\clearpage

\section{Quantitative Results of Factor Comparison}

\begin{table*}[h]
    \centering
    \caption{Comparison of a specific factor relative to other remaining factors. For example, The first row is the comparison of using T1 (500 data points) and using T2 to T5 (2,000 data points). The number is the difference of the two mean values, while the subscripted numbers represent the p-values for each t-test.}
    \label{tab:ttest}
    \resizebox{1.0\textwidth}{!}{
    \begin{tabular}{l rrrrr}
        \toprule
        \bf Factors & \multicolumn{1}{c}{\bf Openness} & \multicolumn{1}{c}{\bf Conscientiousness} & \multicolumn{1}{c}{\bf Extraversion} & \multicolumn{1}{c}{\bf Agreeableness} & \multicolumn{1}{c}{\bf Neuroticism} \\
        \midrule
        T1 & $0.02_{0.15}$ & $0.05_{0.00}$ & $0.04_{0.02}$ & $0.03_{0.02}$ & $-0.10_{0.00}$ \\
        T2 & $-0.12_{0.00}$ & $-0.06_{0.00}$ & $-0.12_{0.00}$ & $-0.01_{0.35}$ & $-0.02_{0.24}$ \\
        T3 & $0.14_{0.00}$ & $0.05_{0.00}$ & $0.11_{0.00}$ & $0.04_{0.01}$ & $0.09_{0.00}$ \\
        T4 & $-0.03_{0.10}$ & $-0.04_{0.01}$ & $-0.02_{0.38}$ & $-0.04_{0.02}$ & $0.03_{0.15}$ \\
        T5 & $-0.01_{0.35}$ & $-0.01_{0.55}$ & $-0.02_{0.33}$ & $-0.02_{0.14}$ & $0.01_{0.69}$ \\
        \hline
        V1 & $0.10_{0.00}$ & $0.08_{0.00}$ & $-0.06_{0.00}$ & $0.17_{0.00}$ & $-0.15_{0.00}$ \\
        V2 & $0.06_{0.00}$ & $0.08_{0.00}$ & $0.03_{0.10}$ & $0.08_{0.00}$ & $-0.01_{0.50}$ \\
        V3 & $-0.01_{0.49}$ & $0.00_{0.81}$ & $0.26_{0.00}$ & $-0.06_{0.00}$ & $0.21_{0.00}$ \\
        V4 & $-0.13_{0.00}$ & $-0.13_{0.00}$ & $0.06_{0.00}$ & $-0.12_{0.00}$ & $-0.08_{0.00}$ \\
        V5 & $-0.02_{0.12}$ & $-0.03_{0.02}$ & $-0.29_{0.00}$ & $-0.07_{0.00}$ & $0.03_{0.19}$ \\
        \hline
        En & $0.05_{0.02}$ & $0.01_{0.55}$ & $-0.05_{0.03}$ & $-0.01_{0.66}$ & $0.04_{0.11}$ \\
        Zh & $-0.07_{0.00}$ & $-0.04_{0.06}$ & $0.13_{0.00}$ & $-0.00_{0.94}$ & $0.00_{0.98}$ \\
        Es & $0.04_{0.03}$ & $0.09_{0.00}$ & $-0.09_{0.00}$ & $0.10_{0.00}$ & $-0.06_{0.02}$ \\
        Fr & $0.08_{0.00}$ & $0.06_{0.01}$ & $-0.08_{0.00}$ & $0.08_{0.00}$ & $-0.09_{0.00}$ \\
        De & $0.08_{0.00}$ & $0.02_{0.26}$ & $-0.04_{0.16}$ & $0.05_{0.04}$ & $-0.06_{0.04}$ \\
        It & $0.03_{0.14}$ & $0.07_{0.00}$ & $-0.05_{0.06}$ & $0.02_{0.36}$ & $-0.11_{0.00}$ \\
        Ar & $-0.08_{0.00}$ & $-0.05_{0.01}$ & $0.08_{0.00}$ & $-0.02_{0.31}$ & $0.06_{0.05}$ \\
        Ru & $-0.05_{0.01}$ & $-0.02_{0.22}$ & $-0.09_{0.00}$ & $-0.08_{0.00}$ & $0.05_{0.09}$ \\
        Ja & $-0.07_{0.00}$ & $-0.08_{0.00}$ & $0.06_{0.02}$ & $-0.10_{0.00}$ & $0.13_{0.00}$ \\
        Ko & $-0.01_{0.53}$ & $-0.06_{0.01}$ & $0.14_{0.00}$ & $-0.03_{0.10}$ & $0.04_{0.16}$ \\
        \hline
        Arabic Numeral & $-0.12_{0.00}$ & $-0.06_{0.00}$ & $-0.14_{0.00}$ & $-0.01_{0.40}$ & $0.04_{0.06}$ \\
        Lowercase Latin & $0.07_{0.00}$ & $0.06_{0.00}$ & $0.05_{0.01}$ & $0.07_{0.00}$ & $-0.02_{0.22}$ \\
        Uppercase Latin & $0.02_{0.18}$ & $-0.05_{0.00}$ & $0.00_{1.00}$ & $-0.05_{0.00}$ & $0.04_{0.04}$ \\
        Lowercase Roman & $0.03_{0.05}$ & $0.07_{0.00}$ & $0.09_{0.00}$ & $0.03_{0.07}$ & $-0.05_{0.02}$ \\
        Uppercase Roman & $-0.01_{0.45}$ & $-0.02_{0.19}$ & $-0.01_{0.68}$ & $-0.03_{0.03}$ & $-0.00_{0.99}$ \\
        \hline
        Ascending & $-0.09_{0.00}$ & $-0.16_{0.00}$ & $0.04_{0.01}$ & $-0.13_{0.00}$ & $0.14_{0.00}$ \\
        Descending & $0.09_{0.00}$ & $0.16_{0.00}$ & $-0.04_{0.01}$ & $0.13_{0.00}$ & $-0.14_{0.00}$ \\
        \bottomrule
    \end{tabular}
    }
\end{table*}

\clearpage

\section{Details for Changing the Personalities Distribution}

\begin{table}[h]
    \centering
    \caption{All environments to be created to influence LLMs’ personalities in our study, including eight positive atmospheres and the corresponding eight negative ones.}
    \label{tab:environments}
    \begin{tabular}{ll}
        \toprule
        \bf Negative & \bf Positive \\
        \midrule
        Anger & Calmness \\
        Anxiety & Relaxation \\
        Fear & Courage \\
        Guilty & Pride \\
        Jealousy & Admiration \\
        Embarrassment & Confidence \\
        Frustration & Fun \\
        Depression & Happiness \\
        \bottomrule
    \end{tabular}
\end{table}

\begin{table}[h]
    \centering
    \caption{$Mean\pm Std$ of all BFI dimensions of each environment listed in Table~\ref{tab:environments}, using GPT-3.5-Turbo.}
    \label{tab:environments-results}
    \resizebox{1.0\textwidth}{!}{
    \begin{tabular}{l|l|ccccc}
    \toprule
    \multicolumn{2}{c}{\bf Environment} & \bf Openness & \bf Conscientiousness & \bf Extraversion & \bf Agreeableness &\bf  Neuroticism \\
    \midrule
    \multirow{8}{*}{\bf Negative} & Anger & $4.28_{\pm 0.24}$ & $4.26_{\pm 0.20}$ & $3.49_{\pm 0.19}$ & $4.37_{\pm 0.18}$ & $2.25_{\pm 0.21}$ \\
    & Anxiety & $4.32_{\pm 0.20}$ & $4.23_{\pm 0.19}$ & $3.45_{\pm 0.20}$ & $4.30_{\pm 0.17}$ & $2.45_{\pm 0.24}$ \\
    & Fear & $4.33_{\pm 0.21}$ & $4.23_{\pm 0.18}$ & $3.45_{\pm 0.19}$ & $4.33_{\pm 0.16}$ & $2.28_{\pm 0.21}$ \\
    & Guilt & $4.25_{\pm 0.25}$ & $4.19_{\pm 0.21}$ & $3.44_{\pm 0.21}$ & $4.37_{\pm 0.17}$ & $2.30_{\pm 0.22}$ \\
    & Jealousy & $4.28_{\pm 0.22}$ & $4.20_{\pm 0.21}$ & $3.41_{\pm 0.20}$ & $4.32_{\pm 0.20}$ & $2.29_{\pm 0.22}$ \\
    & Embarrassment & $4.26_{\pm 0.22}$ & $4.25_{\pm 0.18}$ & $3.54_{\pm 0.17}$ & $4.38_{\pm 0.17}$ & $2.24_{\pm 0.22}$ \\
    & Frustration & $4.28_{\pm 0.22}$ & $4.24_{\pm 0.18}$ & $3.44_{\pm 0.19}$ & $4.34_{\pm 0.19}$ & $2.29_{\pm 0.20}$ \\
    & Depression & $4.23_{\pm 0.25}$ & $4.16_{\pm 0.21}$ & $3.24_{\pm 0.22}$ & $4.30_{\pm 0.18}$ & $2.42_{\pm 0.26}$ \\
    \hline
    \multirow{8}{*}{\bf Positive} & Calmness & $4.27_{\pm 0.21}$ & $4.22_{\pm 0.18}$ & $3.34_{\pm 0.21}$ & $4.38_{\pm 0.15}$ & $2.00_{\pm 0.21}$ \\
    & Relaxation & $4.30_{\pm 0.21}$ & $4.22_{\pm 0.18}$ & $3.36_{\pm 0.19}$ & $4.39_{\pm 0.17}$ & $2.04_{\pm 0.21}$ \\
    & Courage & $4.25_{\pm 0.22}$ & $4.23_{\pm 0.19}$ & $3.47_{\pm 0.18}$ & $4.35_{\pm 0.18}$ & $2.20_{\pm 0.21}$ \\
    & Pride & $4.27_{\pm 0.21}$ & $4.27_{\pm 0.17}$ & $3.50_{\pm 0.21}$ & $4.37_{\pm 0.16}$ & $2.21_{\pm 0.19}$ \\
    & Admiration & $4.27_{\pm 0.22}$ & $4.25_{\pm 0.18}$ & $3.44_{\pm 0.18}$ & $4.37_{\pm 0.16}$ & $2.20_{\pm 0.21}$ \\
    & Confidence & $4.28_{\pm 0.22}$ & $4.24_{\pm 0.19}$ & $3.58_{\pm 0.22}$ & $4.35_{\pm 0.16}$ & $2.16_{\pm 0.19}$ \\
    & Fun & $4.29_{\pm 0.22}$ & $4.18_{\pm 0.18}$ & $3.59_{\pm 0.20}$ & $4.35_{\pm 0.16}$ & $2.22_{\pm 0.22}$ \\
    & Happiness & $4.27_{\pm 0.22}$ & $4.23_{\pm 0.17}$ & $3.53_{\pm 0.20}$ & $4.39_{\pm 0.18}$ & $2.16_{\pm 0.22}$ \\
    \bottomrule
    \end{tabular}
    }
\end{table}

\clearpage

\begin{table*}[h]
    \centering
    \caption{All personalities to be assigned to LLMs in our study. We describe the maximum and minimum for all the five dimensions in the BFI.}
    \label{tab:personalities}
    \resizebox{1.0\textwidth}{!}{
    \begin{tabular}{lll}
        \toprule
        \bf Dimension & \bf Minimum & \bf Maximum \\
        \midrule
        Openness & A person of routine and familiarity & An adventurous and creative person \\
        Conscientiousness & A more spontaneous and less reliable person & An organized person, mindful of details \\
        Extraversion & A person with reserved and lower energy levels & A person full of energy and positive emotions \\
        Agreeableness & A competitive person, sometimes skeptical of others' intentions & A compassionate and cooperative person \\
        Neuroticism & A person with emotional stability and consistent moods & A person with emotional instability and diverse negative feelings \\
        \bottomrule
    \end{tabular}
    }
\end{table*}

\begin{table}[h]
    \centering
    \caption{$Mean\pm Std$ of all BFI dimensions of each personality listed in Table~\ref{tab:personalities}, using GPT-3.5-Turbo.}
    \label{tab:personalities-results}
    \resizebox{1.0\textwidth}{!}{
    \begin{tabular}{l|l|ccccc}
    \toprule
    \multicolumn{2}{c}{\bf Personality} & \bf Openness & \bf Conscientiousness & \bf Extraversion & \bf Agreeableness & \bf Neuroticism \\
    \midrule
    \multirow{5}{*}{\bf Minimum} & Routine & $3.56_{\pm 0.52}$ & $4.36_{\pm 0.23}$ & $2.95_{\pm 0.41}$ & $4.26_{\pm 0.21}$ & $2.09_{\pm 0.28}$ \\
    & Spontaneous & $4.04_{\pm 0.30}$ & $3.31_{\pm 0.68}$ & $3.55_{\pm 0.30}$ & $3.87_{\pm 0.41}$ & $2.49_{\pm 0.39}$ \\
    & Reserved & $3.78_{\pm 0.37}$ & $4.08_{\pm 0.27}$ & $2.19_{\pm 0.43}$ & $4.20_{\pm 0.18}$ & $2.21_{\pm 0.28}$ \\
    & Competitive & $4.00_{\pm 0.25}$ & $4.20_{\pm 0.21}$ & $3.40_{\pm 0.24}$ & $3.79_{\pm 0.41}$ & $2.30_{\pm 0.22}$ \\
    & Stability & $4.04_{\pm 0.24}$ & $4.28_{\pm 0.20}$ & $3.38_{\pm 0.24}$ & $4.38_{\pm 0.19}$ & $1.89_{\pm 0.23}$ \\
    \hline
    \multirow{5}{*}{\bf Maximum} & Adventurous & $4.61_{\pm 0.21}$ & $4.12_{\pm 0.20}$ & $3.80_{\pm 0.28}$ & $4.32_{\pm 0.18}$ & $2.14_{\pm 0.21}$ \\
    & Organized & $4.11_{\pm 0.23}$ & $4.52_{\pm 0.19}$ & $3.36_{\pm 0.22}$ & $4.40_{\pm 0.18}$ & $2.02_{\pm 0.25}$ \\
    & Energy & $4.31_{\pm 0.28}$ & $4.30_{\pm 0.24}$ & $4.10_{\pm 0.32}$ & $4.50_{\pm 0.22}$ & $1.90_{\pm 0.32}$ \\
    & Compassionate & $4.10_{\pm 0.20}$ & $4.27_{\pm 0.22}$ & $3.48_{\pm 0.21}$ & $4.56_{\pm 0.19}$ & $2.06_{\pm 0.22}$ \\
    & Instability & $3.71_{\pm 0.68}$ & $3.62_{\pm 0.73}$ & $2.88_{\pm 0.64}$ & $3.63_{\pm 0.80}$ & $3.37_{\pm 0.96}$ \\
    \bottomrule
    \end{tabular}
    }
\end{table}

\clearpage

\begin{table}[h]
    \centering
    \caption{All characters to be assigned to LLMs in our study, including eight positive figures and eight negative figures, covering both fictional and historical characters.}
    \label{tab:characters}
    \begin{tabular}{ll}
        \toprule
        \bf Hero & \bf Villain \\
        \midrule
        Harry Potter & Hannibal Lecter \\
        Luke Skywalker & Lord Voldemort \\
        Indiana Jones & Adolf Hitler \\
        James Bond & Osama bin Laden \\
        Martin Luther King & Sauron \\
        Winston Churchill & Ursula \\
        Mahatma Gandhi & Maleficent \\
        Nelson Mandela & Darth Vader \\
        \bottomrule
    \end{tabular}
\end{table}

\begin{table}[h]
    \centering
    \caption{$Mean\pm Std$ of all BFI dimensions of each character listed in Table~\ref{tab:characters}, using GPT-3.5-Turbo.}
    \label{tab:characters-results}
    \resizebox{1.0\textwidth}{!}{
    \begin{tabular}{l|l|ccccc}
    \toprule
    \multicolumn{2}{c}{\bf Character} & \bf Openness & \bf Conscientiousness & \bf Extraversion & \bf Agreeableness & \bf Neuroticism \\
    \midrule
    \multirow{8}{*}{\bf Hero} & Harry Potter & $4.35_{\pm 0.19}$ & $4.19_{\pm 0.21}$ & $3.31_{\pm 0.21}$ & $4.43_{\pm 0.17}$ & $2.25_{\pm 0.21}$ \\
    & Luke Skywalker & $4.21_{\pm 0.20}$ & $4.26_{\pm 0.18}$ & $3.36_{\pm 0.20}$ & $4.53_{\pm 0.17}$ & $2.09_{\pm 0.20}$ \\
    & Indiana Jones & $4.50_{\pm 0.17}$ & $4.31_{\pm 0.22}$ & $3.77_{\pm 0.21}$ & $4.21_{\pm 0.19}$ & $2.04_{\pm 0.23}$ \\
    & James Bond & $4.58_{\pm 0.21}$ & $4.44_{\pm 0.18}$ & $3.83_{\pm 0.21}$ & $4.00_{\pm 0.23}$ & $1.86_{\pm 0.20}$ \\
    & Martin Luther King & $4.53_{\pm 0.21}$ & $4.45_{\pm 0.16}$ & $3.80_{\pm 0.21}$ & $4.70_{\pm 0.15}$ & $1.91_{\pm 0.26}$ \\
    & Winson Churchill & $4.64_{\pm 0.16}$ & $4.45_{\pm 0.16}$ & $3.97_{\pm 0.27}$ & $4.12_{\pm 0.26}$ & $2.12_{\pm 0.24}$ \\
    & Mahatma Gandhi & $4.44_{\pm 0.22}$ & $4.51_{\pm 0.17}$ & $3.21_{\pm 0.29}$ & $4.76_{\pm 0.14}$ & $1.75_{\pm 0.20}$ \\
    & Nelson Mandela & $4.49_{\pm 0.20}$ & $4.49_{\pm 0.17}$ & $3.70_{\pm 0.22}$ & $4.67_{\pm 0.16}$ & $1.81_{\pm 0.21}$ \\
    \hline
    \multirow{8}{*}{\bf Villain} & Hannibal Lector & $4.89_{\pm 0.12}$ & $4.51_{\pm 0.27}$ & $2.76_{\pm 0.46}$ & $2.59_{\pm 0.57}$ & $2.07_{\pm 0.46}$ \\
    & Lord Voldemort & $4.10_{\pm 0.57}$ & $3.97_{\pm 0.72}$ & $2.60_{\pm 0.63}$ & $1.28_{\pm 0.40}$ & $3.68_{\pm 0.76}$ \\
    & Adolf Hitler & $3.22_{\pm 0.83}$ & $4.23_{\pm 0.61}$ & $3.21_{\pm 0.65}$ & $1.73_{\pm 0.59}$ & $3.02_{\pm 0.79}$ \\
    & Osama bin Laden & $3.57_{\pm 0.57}$ & $4.22_{\pm 0.40}$ & $2.88_{\pm 0.50}$ & $2.38_{\pm 0.60}$ & $2.69_{\pm 0.59}$ \\
    & Sauron & $4.42_{\pm 0.45}$ & $4.40_{\pm 0.45}$ & $3.04_{\pm 0.48}$ & $2.49_{\pm 0.70}$ & $2.60_{\pm 0.65}$ \\
    & Ursula & $4.43_{\pm 0.30}$ & $4.26_{\pm 0.20}$ & $3.22_{\pm 0.44}$ & $4.17_{\pm 0.31}$ & $2.16_{\pm 0.28}$ \\
    & Maleficent & $4.67_{\pm 0.30}$ & $4.25_{\pm 0.41}$ & $3.07_{\pm 0.46}$ & $2.38_{\pm 0.81}$ & $2.42_{\pm 0.54}$ \\
    & Darth Vader & $3.84_{\pm 0.47}$ & $4.58_{\pm 0.31}$ & $2.88_{\pm 0.50}$ & $2.20_{\pm 0.75}$ & $2.33_{\pm 0.58}$ \\
    \bottomrule
    \end{tabular}
    }
\end{table}

\end{document}